\documentclass{ifacconf}

\usepackage{graphicx}      
\usepackage{natbib}        
\usepackage[export]{adjustbox}
\usepackage{nameref}
\usepackage{amsmath}
\usepackage{latexsym}

\usepackage{xurl}
\usepackage{booktabs}
\usepackage{amssymb}

\usepackage[ruled,vlined,algo2e]{algorithm2e}
\usepackage{setspace}
\usepackage[dvipsnames]{xcolor}
\usepackage{changes}

\begin{document}
\begin{frontmatter}

\title {Temporal Logic Guided Safe Navigation for Autonomous Vehicles}


\author{Aditya Parameshwaran,} 
\author{Yue Wang}

\address{Dept. of Mechanical Eng., Clemson University, Clemson, SC 29634 (e-mails: \{aparame, yue6\}@clemson.edu)}

\begin{abstract}                
Safety verification for autonomous vehicles (AVs) and ground robots is crucial for ensuring reliable operation given their uncertain environments. Formal language tools provide a robust and sound method to verify safety rules for such complex cyber-physical systems. In this paper, we propose a hybrid approach that combines the strengths of formal verification languages like Linear Temporal Logic (LTL) and Signal Temporal Logic (STL) to generate safe trajectories and optimal control inputs for autonomous vehicle navigation. We implement a symbolic path planning approach using LTL to generate a formally safe reference trajectory. A mixed integer linear programming (MILP) solver is then used on this reference trajectory to solve for the control inputs while satisfying the state, control and safety constraints described by STL. We test our proposed solution on two environments and compare the results with popular path planning algorithms. In contrast to conventional path planning algorithms, our formally safe solution excels in handling complex specification scenarios while ensuring both safety and comparable computation times.
\end{abstract}

\begin{keyword}
Safety, Temporal logic, Optimal Control Theory, Unmanned Ground and Aerial Vehicles, Trajectory Tracking and Path Following
\end{keyword}

\end{frontmatter}

\section{Introduction}
Safety verification techniques for autonomous vehicles (AVs) and robots has seen an unprecedented rise in demand over the past few years. Designing the correct set of rules for such complex systems is essential in maintaining the vehicle's safety in the case of uncertainties. Temporal logic offers a formal method to help designers transform natural language rules for safety into mathematical formulae. Temporal operators have been used to express safety requirements and reach-avoid sets with the ability to time and sequence them. Formal verification is one of the most efficient methods to validate design specifications. With the help of model checking, we formally evaluate the functional properties of the system against critical design specifications to rectify any property violations [\cite{rao2023formal}]. Several tools like NuSMV, UPPAL, PAT, and SPIN [\cite{meenakshi2006tool}] can achieve this. NuSMV is a satisfiability (SAT) based model checker to efficiently verify system properties by encoding them into propositional logic formulas, which are then checked for satisfiability using efficient SAT solvers.

Formal specification languages like linear temporal logic (LTL) [\cite{kloetzer2008fully} \& \cite{alur2000discrete}], Metric Temporal Logic (MTL) [\cite{koymans1990specifying} \& \cite{fainekos2006robustness}], signal temporal logic (STL) [\cite{lindemann2017robust} \& \cite{saha2016milp}] have been used to verify and synthesize controllers for motion planning of dynamic autonomous systems that satisfy the predefined specifications. LTL tasks exhibit versatile applicability in standard navigation scenarios [\cite{fainekos2009temporal}, \cite{ding2014optimal}, \cite{guo2015multiagent}, \cite{ulusoy2012robust}], however this leads to a highly complex transition system model based on discrete states that are difficult to generate and can lead to state space explosion [\cite{baier2008principles}]. Our previous work on controllers based purely on STL [\cite{parameshwaran2023safety}] performs better for linearized systems but can have the disadvantage of high computation times due to a larger feasible set. In this work, we combine LTL and STL-based methods to synthesize robot trajectories that maintain formal guarantees and reduce computation loads compared to our previous work.

Our work aims to generate formally safe trajectories and the optimal control inputs required to follow those trajectories using a mixed integer linear programming (MILP) solver. The novelty of our work is that we generate reference trajectories for the MILP solver using an SAT-based model checker based on the concept of counterexample path planning. This ensures the safety and liveness guarantees of the temporal logic specification $\phi$ during our path planning and control sequences. The constraints on the MILP solver ensure minimum safety robustness values $\rho_{min}(\phi)$ while calculating the optimal control inputs. The remainder of the paper is as follows: Section \ref{sec:kinematic_model} describes the AV kinematics model and introduces STL, LTL and model checking. Section \ref{problem} discusses the problem formulation, with feedback linearization, MILP formulation and the algorithm for the solution. In Section \ref{sec:Simulations}, we test our algorithm on two 2D environments and discuss the results in Section \ref{sec:discussion}. The paper concludes with Section \ref{sec:conclussion} and the future work.

\section{Preliminaries}\label{sec:kinematic_model}
\subsection{Nonlinear AV Kinematics Model} Similar to our previous work [\cite{parameshwaran2023safety}], we consider a simplified 2D unicycle kinematic model for an AV where:
\begin{equation} \label{eq1}
    \begin{aligned}
    \left[\begin{array}{c}
    \dot{x} \\
    \dot{y} \\
    \dot{\theta}\\
    \end{array}\right]=\left[\begin{array}{c}
    v \cos \theta \\
    v  \sin \theta \\
    \omega \\
    \end{array}\right] 
    \end{aligned}
\end{equation}
The 2D pose and orientation of the AV are represented by the states $[x, y, \theta]^T$, while $[v, \omega]^T$ denote the AV's linear velocity and angular velocity, respectively. A significant challenge in many nonlinear models is the real-time computation of solutions for non-convex optimization problems [\cite{liu2017real}]. MILP solvers can be used for solving such complex optimization problems with both linear and discrete decision variables. However,  MILP solvers can have longer computational times when combined with multiple state constraints and linear equations. Hence, we use a simple kinematic model that we linearize using feedback linearization in Section \ref{feedbacklin} to decrease the computation times and test out our navigation solution.

\subsection{Signal Temporal Logic (STL)} \label{subsec:stl}
STL allows designers to formulate natural language rules using mathematical operators. This allows for a formal method to set constraints on signal-based systems called specifications. STL has a library of operators representing the formal language's grammar and alphabet, as shown below.

\subsubsection{STL Operators [\cite{donze2013signal}]:}
Writing in the Backus-Naur form, an STL formula can be expressed with the following syntax:
\begin{equation*}
\phi ::= \top\ | \ \mu\ | \ \neg\phi \ | \ \phi \land \psi \ | \ \Box_{[a,b]} \phi\ | \ \Diamond_{[a,b]} \phi\ |\ \phi\ \mathcal{U}_{[a,b]}\ \psi   
\end{equation*}

where $\top$ is the Boolean operator true, $\mu$ is a predicate of the form $f(s(t)) > 0$, 
$\phi$ and $\psi$ are STL specifications, and $s(t)$ is a signal concerning which we check the specifications at time $t$. The Boolean operators $\neg$ signifies negation, and $\land$ and $\lor$ denotes conjunction and disjunction, respectively. Whereas the temporal operators, such as $\mathcal{U}$, $\Box$, and $\Diamond$, signify the `Until,' `Always,' and `Eventually' operators. For a given signal $s(t)$, the list of temporal operators can be defined as follows: 
\begin{equation}
\begin{array}{lll}

s(t) \models \mu & \Leftrightarrow & f\left(s(t)\right)>0  \\
s(t) \models \neg \phi & \Leftrightarrow & \neg\left(s(t) \models \phi\right) \\
s(t) \models \phi \wedge \psi & \Leftrightarrow & \left(s(t) \models \phi\right) \wedge\left(s(t) \models \psi\right) \\
s(t) \models \phi \vee \psi & \Leftrightarrow & \neg(\neg\left(s(t) \models \phi\right) \wedge \neg\left(s(t) \models \psi\right)) \\
s(t) \models \phi \Rightarrow \psi & \Leftrightarrow & \neg\left(s(t) \models \phi\right) \vee\left(s(t) \models \psi\right) \\
s(t) \models \Diamond_{[a, b]} \phi & \Leftrightarrow & \exists t^{\prime} \in[t+a, t+b] \text { s.t. }  s(t^{\prime})\models \phi \\
s(t) \models \square_{[a, b]} \phi & \Leftrightarrow & \forall t^{\prime} \in[t+a, t+b] \text { s.t. } s(t^{\prime}) \models \phi \\
s(t) \models \phi \ \mathcal{U}_{[a, b]} \psi & \Leftrightarrow & \exists t^{\prime} \in[t+a, t+b] \text { s.t. }\left(s(t^{\prime}) \models \psi\right) \\
& & \wedge\left(s(0) \models \square_{\left(0, t \right]} \phi\right)
\end{array}\nonumber
\end{equation}
The expression $\phi$ $\mathcal{U}{[a, b]} \psi$ denotes that within the time interval $t' = [t+a,t+b]$, the proposition $\psi$ will eventually hold if the proposition $\phi$ has been continuously satisfied for time $(0,t]$. Here, the terms `Eventually' and `Always' signify that the propositions must hold \textbf{TRUE} at least once and persist throughout the specified period, respectively. These operators can be consistently formulated using literals in their \textit{Negation Normal Form}, as demonstrated by \cite{lavalle2006planning}. They are not limited in scope and can be applied to any STL operators. Specifications can be amalgamated to formulate constraints for AV models necessitating control over multiple parameters. The signal $s(t)$ can represent any real-time state value such as linear velocity ($v$), position coordinates $[x,y]^T$, etc.

\subsubsection{Robustness Semantics:}

To rigorously assess the fulfillment of an STL specification, robust semantics are employed to delineate real-valued functions denoted as $\rho^\phi$. The robustness metric for any specification $\phi$ within a system, represented by a signal $s$ over the time span $t$, is articulated as follows:

\begin{equation}\label{eq7}
s(t) \models \phi \leftrightarrow \rho^\phi(s(t)) > 0
\end{equation}

Here, $\rho^\phi$ signifies the robustness metric for the given specification at time $t$, wherein the value of this metric may dynamically change over time, specifically during each instant within the interval $t \in [a,b]$ where the specification $\phi_{[a,b]}$ remains valid. The underlying concept behind the robustness metric is to furnish an optimization and control parameter capable of quantifying the extent of satisfaction with the specification. A positive value of robustness ($\rho^\phi > 0$) ensures the fulfillment of the specification throughout the designated time frame.


\subsubsection{\textbf{Example:}}
Suppose there is an STL specification demanding that $x(t) \geq 1$ `Always' holds for all time instances. We define the predicate $\mu(t) = x(t) - 1$, ensuring $(\mu(t) \geq 0)$ holds for all $t \in [0, \infty]$. The STL specification alongside the safety robustness measure $\rho^{\phi}$ is as follows:
\begin{equation}
\phi = {\square_{[0,\infty]}
(\mu(t) \geq 0)}\nonumber
\label{eqn:eq7}
\end{equation}
\begin{equation}
\rho^{\phi}(x(t)) = \mu(t)
\label{eqn:eq8}
\end{equation}

To justify the complete satisfaction of the above specification for all given times, we imply the method of quantifying the specification value concerning the predicate being measured as a form of the robustness ($\rho$) of the signal.

\subsection{Linear Temporal Logic (LTL)}\label{subsubsec:ltl}
LTL is another formal language used to specify rules on system functionality and properties. While STL is powerful for continuous systems and can express a wider range of temporal properties involving both time and signal values, LTL's primary advantage is its effectiveness and efficiency in the context of synchronous, discrete-time systems. This means that the present moment refers to the current state, and the next moment corresponds to the immediate successor state [\cite{baier2008principles}]. LTL provides a set of operators that allow designers to express temporal relationships and constraints between events in a system. The syntax for an LTL formula $\phi$  can be defined recursively by the following grammar:
\begin{equation*}
\phi ::= \top \ | \ \bot \ | \ a \ | \ \neg\phi \ | \ \phi \land \phi \ | \ \phi \lor \phi \ | \ X\phi \ | \ \phi \ \mathcal{U} \psi
\end{equation*}
Here, $a \in AP$ denotes that $a$ is an atomic proposition in the set of all atomic propositions $(AP)$. In a transition system, $a$ can represent a specific state label or condition that can be true or false at a given state. $\bot$ is the Boolean operator `False' and $X$ denotes the `Next' operator. The rest of the basic operators have the same meaning as in the STL operators in Section \ref{subsec:stl}.
\subsubsection{LTL Model Checking:}\label{modelchecking}
Model checking systematically verifies whether a system satisfies a specified criterion represented by temporal logic. For an LTL specification $\phi$, model checking verifies $\phi$'s satisfiability by checking if there exists a trace $\pi$ such that $\pi \models \psi$, where $\psi = \neg \phi$. If a trace is found that satisfies $\psi$, then that trace violates $\phi$, and the system does not satisfy $\phi$. Typically, it is assumed that $\pi_0 \in S_0$, where $S_0 \subseteq S$ is a designated set of initial states. We use NuSMV to perform model checking, which supports Boolean, scalar, and finite arrays for finite state machine representations. Upon defining the system model, NuSMV exhaustively explores the state space to verify specified properties. If specifications are met, the model checker confirms \textbf{TRUE}; otherwise, NuSMV generates the shortest counterexample of states aiding in error rectification.

\section{Problem Formulation} \label{problem}
\subsection{Feedback Linearized Kinematics Model:} \label{feedbacklin}

To circumvent infeasible solutions and mitigate computational overhead, we propose the utilization of a linear state-space 
system model, wherein:
\begin{equation}\label{eq:linearAVmodel}
    \begin{aligned}
    &\dot{x}_1 = \dot{x} = v_x  = x_3 \\ 
    &\dot{x}_2 = \dot{y} = v_y  = x_4 \\ 
    &\dot{x}_3 = \dot{v}_x = a_x  = u_1 \\ 
    &\dot{x}_4 = \dot{v}_y = a_y  = u_2 \\ 
    \end{aligned}
\end{equation}
In the proposed model (\ref{eq:linearAVmodel}), the states are denoted by 
 $\mathbf{x}(t) \triangleq [x_1(t), x_2(t),x_3(t),x_4(t)]^T= [x(t), y(t), v_x(t),v_y(t)]^T$, representing the global X-Y coordinates and X-Y linear velocities, respectively. The control inputs are defined as $\mathbf{u}(t) = [u_1,u_2]^T = [a_x,a_y]^T$, signifying the linear accelerations in the X and Y coordinates, respectively. 


However, we still need to use the original nonlinear kinematic model (\ref{eq1}) to obtain the robot states $[x,y]^T$ given the optimal control inputs $ [u^*_1,u^*_2]^T $computed by the MILP solver. Hence, we introduce a new set of coordinates and relate Equation (\ref{eq:linearAVmodel}) with Equation (\ref{eq1}) by feedback linearization [\cite{DeLuca2000StabilizationOT}].  By choosing the position $[x,y]^T$ of the AV as the system output and differentiating (\ref{eq1}), we can obtain the following:
\begin{equation}
\label{eq:pos2control}
\left[\begin{array}{l}
\ddot{x} \\
\ddot{y}
\end{array}\right]=\left[\begin{array}{cc}
\cos (\theta) & -\sin (\theta) \\
\sin (\theta) & \cos (\theta)
\end{array}\right]\left[\begin{array}{c}
\dot{v} \\
v \omega
\end{array}\right]
\end{equation}
We can then invert the matrices in Equation (\ref{eq:pos2control}) to calculate $[v,\omega]$ such that:
\begin{equation}\label{eq4:control_input}
\left[\begin{array}{c}
\dot{v} \\
v \omega
\end{array}\right]=\left[\begin{array}{cc}
\cos (\theta) & \sin (\theta) \\
-\sin (\theta) & \cos (\theta)
\end{array}\right]\left[\begin{array}{l}
u_1 \\
u_2
\end{array}\right]
\end{equation}
where $[u_1,u_2]^T = [a_x,a_y]^T$ are the inputs for the linearized model described in \eqref{eq:linearAVmodel}. These inputs are generated from the MILP solver as the optimal solution to the vehicle navigation problem to be presented in Section \ref{sec:MILP}.

We also discretize the system with a time step $T_s$ as the MILP solver works only for discrete-time models. From (\ref{eq:linearAVmodel}), we formulate the linear discrete-time state-space systems as follows:
\begin{equation}\label{eq2}
    \begin{aligned}
    &\mathbf{x}(k+1) = A\mathbf{x}(k) + B\mathbf{u}(k)\\
    &\mathbf{y}(k) = C\mathbf{x}(k)
    \end{aligned}
\end{equation}
where
\begin{equation}\label{eq3}
    \begin{aligned}
    A = \left[\begin{array}{c}
        0 \ 0 \ 1 \ 0  \\
        0 \ 0 \ 0 \ 1   \\
        0 \ 0 \ 0 \ 0   \\
        0 \ 0 \ 0 \ 0
    \end{array}\right] \ ;
 \    B = \left[\begin{array}{c}
        0 \ 0 \  \\
        0 \ 0 \   \\
        1 \ 0 \   \\
        0 \ 1 \
    \end{array}\right]
    \end{aligned} \ ;
 \  C =\left[\begin{array}{c}
        1 \ 0 \ 0 \ 0  \\
        0 \ 1 \ 0 \ 0   \\
        0 \ 0 \ 1 \ 0  \\
        0 \ 0 \ 0 \ 1   \\
    \end{array}\right] \ \nonumber
\end{equation}

\subsection{Symbolic Path Planning:}\label{symbolic}

In our research, we utilize symbolic path planning based on the principle of model checking. We use the NuSMV model checker to check the AV mission environment with the negation of the LTL specification. If a counterexample is identified, this is equivalent to finding a path for the AV that satisfies the LTL specification. A counterexample for a design specification $\phi$ constitutes a valid trace ($\tau$)  of state transitions resulting in the falsification of $\phi$ [\cite{ovsiannikova2020visual}].

\begin{equation}
    \phi(s_1, \ldots, s_n) = \textbf{FALSE} \iff \exists \ \tau(\phi)
\end{equation}

Here, $\phi \in \Phi$ represents a design specification expressed in terms of state variable $s$, while $\tau(\phi) = [s_1, \ldots, s_n]$ denotes the trace of state variable $s$ corresponding to the counterexample of that specification. Multiple traces may falsify a given specification, all formally guaranteed to be safe and compliant with state dynamics, but NuSMV provides the shortest \textit{Trace} possible. Leveraging this trace generation concept, we establish trajectories ensuring the safety of AV navigation.

For symbolic path planning, we formulate an LTL specification as follows:

\begin{equation} \label{eq10}
    \phi = \Diamond(\pi_{goal}) \land \square (\neg \pi_{obstacle})
\end{equation}

where $\pi_{obstacle}$ represents obstacle regions and $\pi_{goal}$ signifies a goal region on the map. As detailed in Section \ref{subsec:stl}, $\Diamond$ and $\square$ denote the `Eventually' and `Always' operators. We equip NuSMV with possible state transitions within a known map environment, allowing the robot to navigate [${\uparrow, \downarrow, \leftarrow, \rightarrow}$] on a grid-based layout. We have noticed that diagonal transitions between cells cause instability issues with the model checker. We also eliminate transitions into obstacle cells and out-of-state limits to yield viable state transition only based on the LTL specification.

\SetKwInOut{Input}{Input}
\SetKwInOut{Output}{Output}
\SetKwInOut{Data}{Data}

\begin{algorithm2e}[!ht]
\setstretch{1.2}

\Input {Goals($\mathbf{x}_{\text{goal}}$), Start($\mathbf{x}_{0}$)}
\Output {\text{Safe Path}($\tau_{\text{MILP}}$)}
\Data {Map($\mathbb{X}_\text{MAP}$), Weight Matrices($Q_1$ , $Q_2$)} 
$goal_{\text{array}} \gets \text{Grid2Array}(\mathbf{x}_{\text{goal}},\mathbb{X}_\text{MAP})$\;
$start_{\text{array}} \gets \text{Grid2Array}
(\mathbf{x}_{0},\mathbb{X}_\text{MAP})$\;
$\text{Get LTL task specification: }\phi = (\Diamond(\pi_{goal}) \land \square (\neg \pi_{obstacle})$\; 
\While{isempty $\tau_{\text{LTL}}$}{
    $\text{Make a NuSMV file using } (\mathbb{X}_\text{MAP}, \mathbf{x}_{\text{goal}},  \mathbf{x}_{0})$\;
    $\text{Symbolic Path Planning using } (\neg \phi)$\;
    $\text{Get terminal output from NuSMV file}$\;
    $\tau_{\text{LTL}} \gets \text{Array2Grid}(\tau) \gets \text{Terminal output}$\;
    }
  
\Input{Constraint matrix($Z$),\\ Initial state($\mathbf{x}_0$)}

\While{$\rho(\tau_{\text{MILP}}) \leq \rho^{\phi}_{min}$}
    {\For{$k = 1$ to $H$}
    {
    $Z \gets [ \ Z, \ \mathbf{x}_{k+1} = A \mathbf{x}_{k} + B \mathbf{u}_{k} \ ] $\;
    $Z \gets [ \ Z, \ \mathbf{y}_{k} = C \mathbf{x}_{k} \ ] $\;
    $Z \gets [ \ Z, \ \mathbf{u}_{min}(:,k) \leq \mathbf{u}_{k} \leq \mathbf{u}_{max}(:,k) \ ]$\;
    $Z \gets [ \ Z,\ \mathbf{x}_{k} \in \mathbb{X}_\text{MAP}]$\;
    $Z \gets [ \ Z,\ \rho_{k} \geq \rho_{min}]$\;
    $\mathbb{J} \gets \mathbb{J} + Q_1  \|\mathbf{u}(k)\|  + Q_2 \|(\mathbf{y}_{k} - \tau_{\text{LTL}})\|$\;
    }
    $\text{Find control inputs: } \mathbf{u}^{*} \gets \ MILP_{solver}(\min (\mathbb{J}))$\;
    $\text{Safe Path (} \tau_{\text{MILP}} ) \gets A\mathbf{x}_{k} + B\mathbf{u}^{*}$\;
    $\rho(\tau_{\text{MILP}}) \gets \text{Robustness Breach: }(\tau_{\text{MILP}}, \phi)$\;
}

\caption{Use NuSMV for symbolic path planning and then use MILP solver for navigation}
\end{algorithm2e}

\subsection{Solution}

 The navigation problem begins by defining the state and control bounds for a known 2D environment and AV. Our solution is based on three main steps:
\begin{enumerate}
  \item A NuSMV file is generated based on the environment and state conditions ($\mathbb{X}_{MAP}$), the goal locations ($\mathbf{x}_{goals}$), the initial robot state ($\mathbf{x}_{0}$) and the LTL specification ($\phi$)
  for symbolic path planning.\\
  \item NuSMV is then used to generate a formally safe trajectory ($\tau_{\text{LTL}}$) for the robot based on the falsification of $\phi$.\\
  \item The counterexample based $\tau_{\text{LTL}}$ is used as a reference trajectory to solve the MILP problem for optimal control inputs ($\mathbf{u}^* =[a^{*}_x,a^{*}_y]^T$), where $[a_x,a_y]^T$ are the linear accelerations as discussed in Section \ref{sec:kinematic_model}.
\end{enumerate}
\begin{figure}[!ht]
      \includegraphics[width=\columnwidth,center]{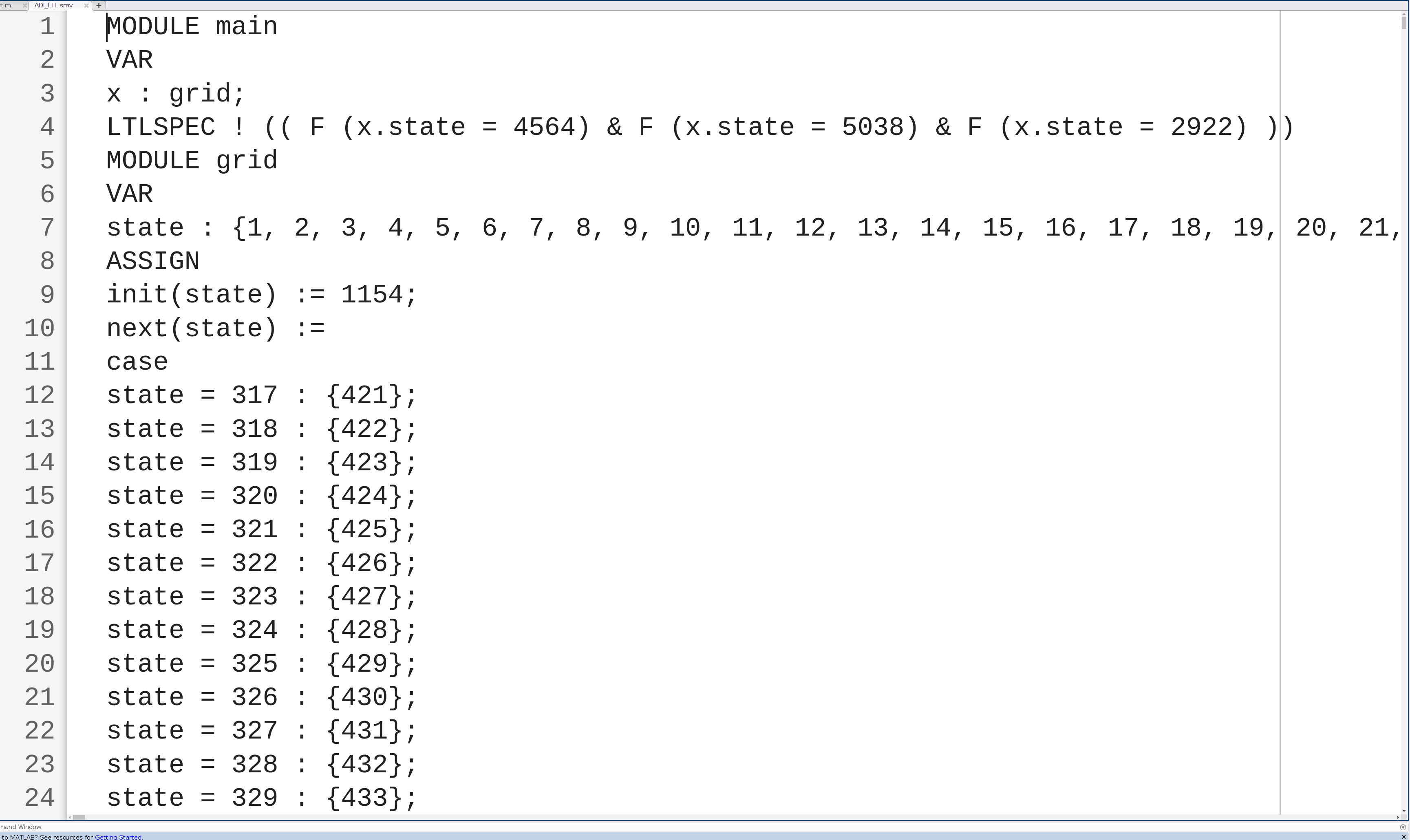}
      \caption{NuSMV file for $\phi_2$}
      \label{fig:NuSMV}
\end{figure}

A snapshot of the NuSMV file is shown in Figure \ref{fig:NuSMV}. Based on the binary grid-based environment, we discretize the map into grid cells that represent the position coordinates $\mathbf{x}_{i,j}$ for the robot in 2D, with free cells as safe states and occupied cells as obstacles for Algorithm 1. 
To generate a counterexample-based path, we modify the specification to $\neg \phi$, prompting the model checker to verify the negation's truth. We specify the robot's initial state as $\mathbf{x}_0$, and if a valid path exists, NuSMV generates a trace of states $\tau = (\mathbf{x}_0, \mathbf{x}_1, \ldots, \mathbf{x}_N)$. This trace is a valid trajectory guiding the robot from $\mathbf{x}_0$ to $\mathbf{x}_N$  following $\phi$ and adheres to state limits. NuSMV ensures finite paths without considering cyclic transitions between grid cells. We generate the reference trajectory $\tau$ from NuSMV output, followed by its conversion to $\tau_{LTL}$. Now, we can use a tracking MILP approach to generate optimal control inputs while following the reference trajectory.

\subsection{Mixed Integer Linear Programming}\label{sec:MILP}
In our previous work [\cite{parameshwaran2023safety}], we used STL-based constraints on our MILP solver to solve for optimal and safe control velocities for our robot. We received optimal control inputs that would satisfy the temporal logic specification in \eqref{eq10} but at a substantially high trade-off with computational efficiency and optimization time. For this work, the proposed solution is to combine symbolic path planning with a reference tracking MILP. This mitigates the problem of high computational time while maintaining formal guarantees of safety. 
The MILP problem can be stated as below:
\begin{equation}\label{eq:MILP}
   \begin{aligned}
    &\mathbb{J} = \min_{\mathbf{u}(k)} \sum_{k=0}^{H-1} \left( \ Q_1 \|\mathbf{u}(k)\|  + Q_2 \|(\mathbf{y}_{k} - \tau_{\text{LTL}})\| \ \right) \\
    & \text{subject to } \ \rho(\tau_{\text{MILP}}) \geq \rho^{\phi}_{\text{min}}, \\
    & \mathbf{x}(k+1) = A\mathbf{x}(k) + B\mathbf{u}(k), \\
    & \mathbf{y}(k) = C\mathbf{x}(k), \\
    & \mathbf{u}(k) \in \mathbb{U}, \ \mathbf{x}(k) \in \mathbb{X}_{MAP}
    \end{aligned} 
\end{equation}

In Algorithm 1, the MILP constraints for the vehicle dynamics, state, control bounds, and STL robustness bounds are added in Lines 11 to 15. After these constraints, the objective function described in Equation \eqref{eq:MILP} is added to the algorithm. The optimal control inputs are calculated by minimizing the cost function while satisfying the constraints. The weight matrices $Q_1$ and $Q_2$ govern the allocation of importance between minimizing the control inputs ($\mathbf{u}^*$) and adhering to the reference trajectory ($\tau_{\text{LTL}}$). Control inputs ($\mathbf{u}^*$) are then converted  to $[v, \omega]^T$ using Equation (\ref{eq4:control_input}), which can be then used to simulate the AV and provide a formally guaranteed optimal path $\tau_{\text{MILP}}$.

\section{Simulations}\label{sec:Simulations}

In this section, we present a comparative analysis of the performance of our LTL + MILP path planner for trajectories within 2D environments. The trajectories obtained from our LTL-based symbolic planner and LTL + MILP solution are systematically evaluated against established AV path planning algorithms, precisely A* and RRT*. To validate our approach, we examine two distinct 2D environments: the Simple and Complex environments, depicted in Figures \ref{fig:simpleMap} and \ref{fig:complexMap}, respectively. 

Simulations are carried out on MATLAB 2023B using {\fontfamily{cmtt}\selectfont Gurobi 11.0} solver through {\fontfamily{cmtt}\selectfont YALMIP} for MILP formulation with 2.9GHz clock speed, eight processors and 16GB of RAM. The feedback linearized model is initialized for a sampling period $T_s = 0.1s$ at different starting locations.

\subsubsection{\textbf{Example A}}
\begin{figure}[!ht]
      \includegraphics[width=\columnwidth,center]{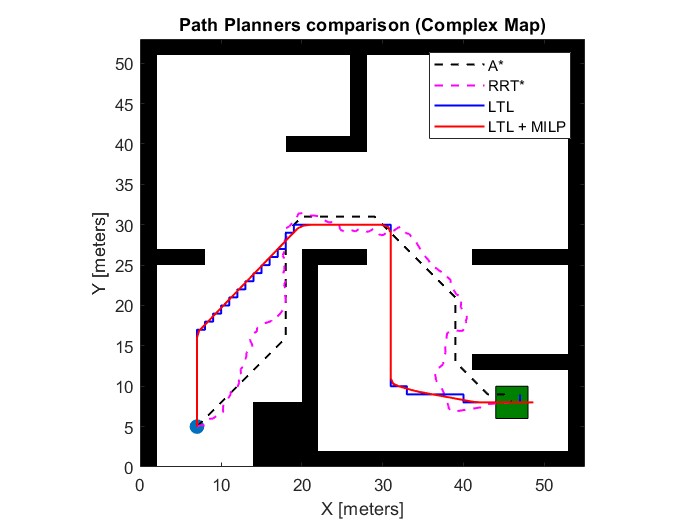}
      \caption{Simple Environment with ($\phi_1$) LTL specification}
      \label{fig:simpleMap}
\end{figure}

:  We consider a single goal approach with LTL specification $\phi = \Diamond(\pi_1) \land 
 \square (\neg \pi_{obstacle})$, where $\pi_{obstacle}$ are considered the wall regions of the environments. The specification states that the robot needs to visit goal location $\pi_1$ while avoiding obstacles regions. The specification assumes that the possible states are viable and that a trajectory satisfies the given specification. In Figure \ref{fig:simpleMap}, we see the results of \textit{\textbf{Example A}} on the Simple Environment with all 4 planners.\\

\subsubsection{\textbf{Example B}}
\begin{figure}[!ht]
      \includegraphics[width=\columnwidth,center]{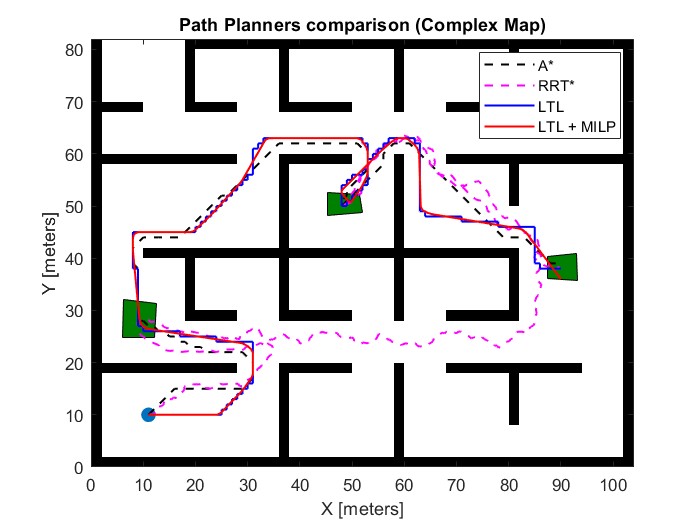}
      \caption{Complex Environment with ($\phi_2$) LTL specification}
      \label{fig:complexMap}
\end{figure}

: Similar to the previous example, we consider a multi-goal approach with LTL specification $\phi = \Diamond(\pi_1 \ \mathcal{U} \ (\pi_2 \ \mathcal{U} \ \pi_3)) \land \square (\neg \pi_{obstacle}) $ for the Complex environment with three goals. The computation times increase as the environment becomes more complex and multiple goals are to be achieved. The A* and RRT* algorithms consecutively search for the optimal path between the goal regions in accordance with the specification. In Figure~\ref{fig:complexMap}, we see the results for \textit{\textbf{Example B}} on the Complex Environment.\\

\section{Discussion}\label{sec:discussion}

In the instances provided, it is evident that both the solutions adhere to the LTL specifications. However, due to the limitation of the symbolic path planner to only four directions of movement, the trajectory $\tau_{LTL}$ is sub-optimal. Conversely, the trajectory $\tau_{\text{MILP}}$ generated by the LTL + MILP approach exhibits greater optimality by referencing $\tau_{LTL}$ while also making corner cuts. A* path-finding yields the most optimal path overall, albeit lacking smoothness in Fig. \ref{fig:simpleMap}. Conversely, RRT* performs the poorest, generating paths that are sub-optimal as they are the longest of the 3.

\begin{figure}[!ht]
      \includegraphics[width=\columnwidth,center]{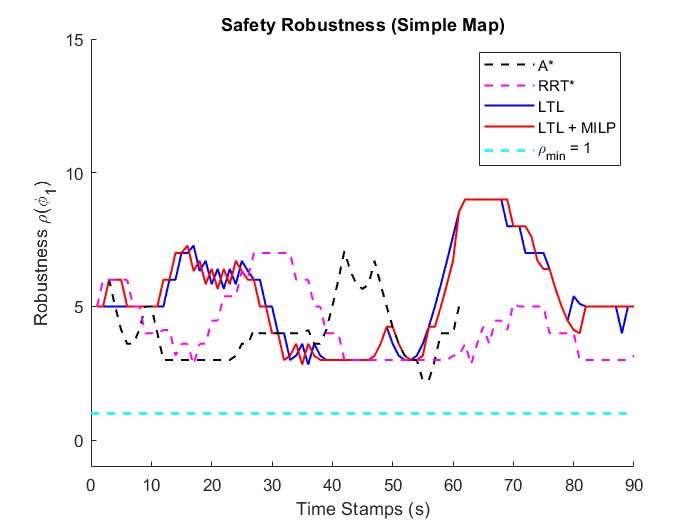}
      \caption{STL Robustness for Simple Environment}
      \label{fig:simpleRho}
\end{figure}

\begin{figure}[!ht]
      \includegraphics[width=\columnwidth,center]{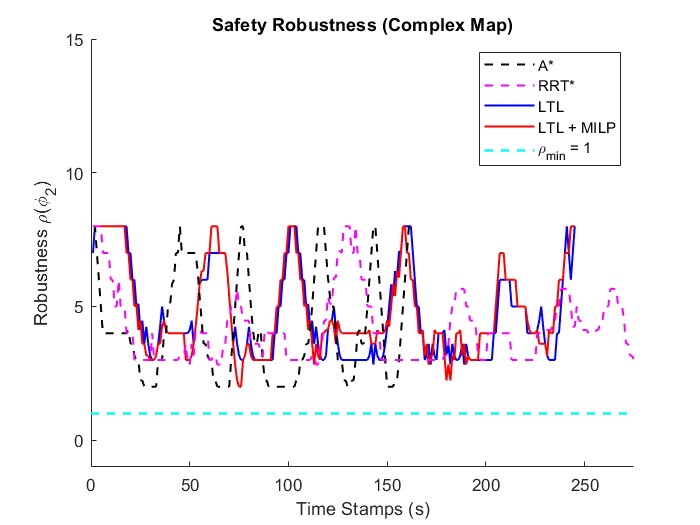}
      \caption{STL Robustness for Complex Environment}
      \label{fig:complexRho}
\end{figure}

We calculate and compare the $\rho(\phi)$ values using the STL model checker {\fontfamily{cmtt}\selectfont Breach} (\cite{donze2010breach}) for $\rho(\phi) \geq \rho_{min}$, where $\rho_{min} = 1$. As it is evident that all path planners reach their respective goals, it implies that {\fontfamily{cmtt}\selectfont Breach} eventually verifies the minimum distance of the paths from the closest wall shown in Fig. \ref{fig:simpleRho} and \ref{fig:complexRho} at any given point in time. While RRT* shows comparable robustness results to our proposed solution, A* exhibits lower $\rho(\phi)$ values as shown in Table \ref{table1}. This indicates that the paths generated by these algorithms are typically less safe than those derived from our approach.


\begin{table}[htpb]
\centering
\caption{Planner comparison}
\begin{adjustbox}{width=0.95\columnwidth}
\begin{tabular}{|l|l|l|l|c|l|l|}
\hline
 & \multicolumn{ 3}{c|}{\textbf{Simple }} & \multicolumn{ 3}{c|}{\textbf{Complex}} \\ \hline
\textbf{Planners} & \multicolumn{1}{c|}{$\rho(\phi_1)$} & \multicolumn{1}{c|}{$T(s)$} & \multicolumn{1}{c|}{$\mathbb{J}$} & $\rho(\phi_2)$ & \multicolumn{1}{c|}{$T(s)$} & \multicolumn{1}{c|}{$\mathbb{J}$} \\ \hline
RRT* & 1.83 & 3.56 & 90 & 1.47 & 4.85 & 611 \\ \hline
A* & 1.00 & 0.99 & 58.8 & 0.96 & 1.47 & 161 \\ \hline
LTL + MILP & 1.83 & 3.01 & 64.9 & 0.96 & 6.92 & 245 \\ \hline
LTL & 1.83 & 1.86 & - & 1.47 & 5.85 &  - \\ \hline
\end{tabular}
\end{adjustbox}
\label{table1}
\end{table}

We see that the A* algorithm works the fastest but leads to moderately safe trajectories as shown by $\rho(\phi)$ values. In comparison, the LTL + MILP approach provides more robust trajectories, with lesser cost ($\mathbb{J}$) and computation times $T(s)$ than RRT*. Additionally, it is essential to note that once our NuSMV transition file is generated, further transitions for the same environment need not be recreated. This means that the computational times drastically decrease for any additional LTL specifications within the same environment.

\section{Conclusion} \label{sec:conclussion}
In this paper, we propose a unique approach to the AV navigation problem using a combination of symbolic path planning and mixed integer linear programming.  We defined a simplified kinematics model for a robot.  Then, we introduced formal verification languages like LTL and STL that can guarantee satisfaction and safety for given specifications.  By leveraging an SAT-based model checker, we generate formally safe trajectories based on the LTL specification while ensuring obstacle avoidance and the achievement of goal locations.  Furthermore, we combine the formally safe trajectories with an MILP solver to formulate a reference trajectory-based optimal control problem.  This approach allowed us to maintain formal satisfaction guarantees while reducing computational overheads compared to our previous methods.  We then demonstrate our efforts in two different environments and LTL specifications while maintaining minimum safety robustness levels for both.  A direction for future work will involve higher dimensional autonomous systems and environments due to STL's capability of handling multiple signals.

\bibliography{ifacconf}

\end{document}